# WEAKLY-SUPERVISED ROI EXTRACTION METHOD BASED ON CONTRASTIVE LEARNING FOR REMOTE SENSING IMAGES

*Lingfeng He[1], Mengze Xu[1], Jie Ma[1*]*

[1]Beijing Foreign Studies University

## ABSTRACT

ROI extraction is an active but challenging task in remote sensing because of the complicated landform, the complex boundaries and the requirement of annotations. Weakly supervised learning (WSL) aims at learning a mapping from input image to pixel-wise prediction under image-wise labels, which can dramatically decrease the labor cost. However, due to the imprecision of labels, the accuracy and time consumption of WSL methods are relatively unsatisfactory. In this paper, we propose a two-step ROI extraction based on contractive learning. Firstly, we present to integrate multiscale Grad-CAM to obtain pseudo pixelwise annotations with well boundaries. Then, to reduce the compact of misjudgments in pseudo annotations, we construct a contrastive learning strategy to encourage the features inside ROI as close as possible and separate background features from foreground features. Comprehensive experiments demonstrate the superiority of our proposal. Code is available at https://github.com/HE-Lingfeng/ROI-Extraction

***Index Terms***— ROI extraction, weakly supervised learning, remote sensing, contrastive learning, deep learning

## 1. INTRODUCTION

Region-of-interest extraction (ROI) aims at obtaining informative regions in remote sensing images (RSI) and has attracted significant attention in the field of remote sensing. It can be applied to multiple scenarios, including urban expansion, environmental change and automatic navigation research. With the rapid development of deep learning technology, especially convolutional neural network, the accuracy and generalization ability have been greatly improved.

Over the last two decades, considering the distinction of supervision mechanism, previous proposals can be divided into three main streams: fully supervised learning methods [1, 2], weakly supervised learning methods [3, 4] and unsupervised learning methods [5, 6].

For fully supervised learning methods, pixel-level annotations are indispensable in training phase. Shelhamer *et al.* [1] proposed the first fully convolutional networks (FCNs) that take an arbitrary size input and produce correspondingly sized output with efficient inference and learning. Ronneberger *et al.* [2] designed the symmetrical encoder-decoder architecture called UNet that captures the context and precise localization information. As for remote sensing images, Chu *et al.* [7] constructed a new network structure Res-UNet by replacing the contraction part of UNet with the residual modules of ResNet which specializes in complicated sea-land segmentation.

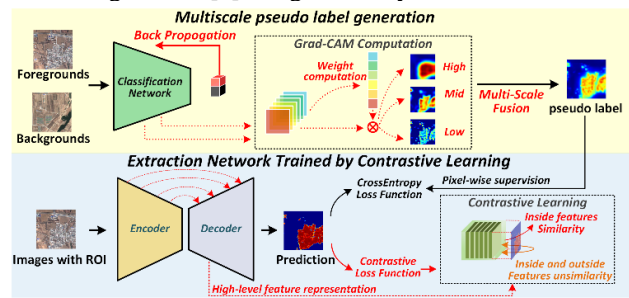

Fig. 1 The architecture of our ROI extraction method based on contrastive learning

As for unsupervised methods, several bottom-up features are designed for unmarked data. Zhang *et al.* [5] proposed a global and local saliency analysis (GLSA) method to obtain texture and edge features of different scales and orientations. Murphy *et al.* [6] presented a novel algorithm consisting of diffusion geometry and partial least squares regression (PLSR) for unsupervised segmentation of hyperspectral imagery.

Compared with fully supervised methods that require pixel-level labels and unsupervised methods with unlabeled data, weakly supervised learning (WSL) prefers image-wise labels, which reflect which or what objects are involved. As for WSL, previous works tend to be two-step approaches, where pseudo pixel-level masks with image-level labels are firstly generated by bottom-up or top-down strategies, and an off-the-shelf network is constructed to learn the mapping between the input images and the pseudo pixel-level masks. Zhang *et al.* [3] adaptively adjusted the contributions of quality-varying pseudo labels and proposed a novel self-paced residual aggregated network (SP-RAN) for solar panel mapping.

As mentioned above, previous WSL methods have achieved remarkable success in ROI extraction. However, as the enhancement of the resolution and urban expansion, the complicated characteristics of RSIs brought new challenges, that the generated pseudo masks with unsatisfied boundary would bring several unpredictable mistakes for later off-the-shell network.

[*] Corresponding author: Jie Ma (majie_sist@bfsu.edu.cn)

In this paper, we propose a ROI extraction method based on contrastive learning for remote sensing images in a weakly supervised manner. First, in order to obtain more accurate pseudo labels, we present to utilize multiscale gradient-weighted class activation map to capture both local, middle and global features. Then, we construct a joint loss function, where the similarity of the features is taken into considerate to eliminate the impact of the mistakes in pseudo labels. Here, loss function is designed to reduce the differences between the high-level features inside ROI and increase the differences between foreground and background high-level features, which introduces more supervision from constraints of feature maps themselves. Experimental results and ablation studies validates the efficiency of our proposal.

## 2. METHODOLOGY

In this work, the proposed WSL method can be divided into two sequential parts including a multiscale pseudo label generation part and an extraction network construction part. The overall framework of our work is illustrated in Figure 1.

### 2.1 Multiscale Pseudo Label Generation

As for classification CNN, the class score is the probability of the input image belonging to corresponding category. To analyze the impact of each pixel in classification task, previous work [8] proposed that the gradient of the class score with respect to the input image indicates the change of which pixels influence the class score most. That is, those salient pixels (pixels with higher intensity) in the gradient map are most likely to be objections of the corresponding class.

Meanwhile, for the multiscale features can enhance the extraction results for a great extent, we decide to compute the gradient with specific convolutional layers to depict elaborate boundaries and suppress complicated backgrounds simultaneously.

Our classification network is inspired by the structure of VGG19. Considering that the middle convolutional layers can extract the high-level semantic features and deeper convolutional layers extract more spatial features, we decide to extract the outputs of the last three layers before the max-pooling layers so that we can aggregate both semantic and spatial features. In our network, these three are 17th, 26th, 35th layers. Then we use these three outputs to generate Grad-CAM respectively and merge them by computing the arithmetical mean. For a given input image, let $A_{x,y}^{k,t}$ represent the activation of the $k_{th}$ feature map in the $t_{th}$ convolutional layer at the spatial location $(x, y)$. According to [8], the Grad-CAM of training images are computed as follows:

$$\alpha_k = \frac{1}{Z}\sum_i\sum_j \frac{\partial y^c}{\partial A_{i,j}^{k,t}} \quad (1)$$

$$S_{Grad-CAM}^t = \text{ReLU}(\sum_k \alpha_k A^{k,t}) \quad (2)$$

$$S_{merged} = \frac{1}{3}\sum_t S_{Grad-CAM}^t \quad (3)$$

where $y^c$ is the class score for class $c$, $t = \{17, 26, 35\}$.

We generate the binary mask $Y$ of $S_{merged}$ with threshold of 0.5 as the pseudo label as shown in Fig. 2. Residential areas in the RSIs are the ROIs in our research.

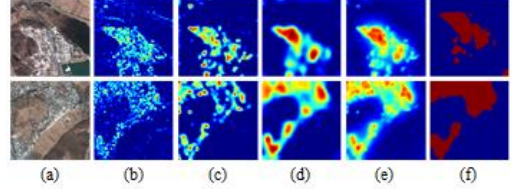

Fig.2 The activation maps (am) are merged to one map: (a) input, (b) am of 17th layer, (c) am of 26th layer, (d) am of 35th layer, (e) merged map, (f) binary mask.

### 2.2 Extraction Network Trained by Contrastive Learning

We integrate UNet with constative learning to train a ROI extraction network utilizing the multiscale pseudo labels obtained in previous part.

Contrastive learning [9] is a popular self-supervised idea and has made great success in recent years. Its goal is to train the network so that the representations of different augmented views of the same classes are as close as possible to each other. Meanwhile, the representation of different views from different instances should be distinctive to each other. Since there are some mislabeled pixels in pseudo annotations, we introduce the contrastive learning to encourage feature and its positive pair to be close in the feature maps and pushing away representations of all other negative pairs. As the convolutional layer goes deeper, features maps will aggregate more spatial features and less semantic features. Therefore, we propose to extract the outputs of the first two upsampling layers in the expanding path, denoted as $up_{1f}$, $up_{2f}$, to conduct contrastive learning algorithm. The detailed process is presented as follow:

Given and input image, the binarized prediction of UNet is denoted as $\tilde{Y}$. We first transform the binarized prediction as the same spatial size of the corresponding features maps. For $up_{1f} \in \mathbb{R}^{H*W*C}$, features in foreground areas as regarded as positive features $k_+ \in \mathbb{R}^{N*C}$, otherwise they will be encoded as negative pairs $queue \in \mathbb{R}^{C*K}$, where $N$ is the number of positive features and $K$ is the number of negative features. Meanwhile, $up_{1f}$ are encoded as $q \in \mathbb{R}^{N*C}$. The positive logits and negative logits are computed as:

$$l_{pos} = q.view(N,1,C) \cdot k_+.view(N,C,1) \quad (4)$$

$$l_{neg} = q.view(N,C) \cdot queue.view(C,K) \quad (5)$$

where $l_{pos} \in \mathbb{R}^{N*1}$, $l_{neg} \in \mathbb{R}^{N*K}$, which measure the similarity between positive features and dissimilarity between positive features and negative features. The logits are concatenated as $logits \in \mathbb{R}^{N*(K+1)}$ to consider both similarity and dissimilarity information.

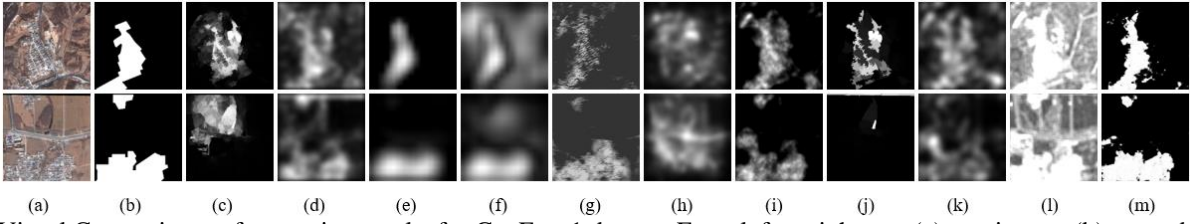

(a)　(b)　(c)　(d)　(e)　(f)　(g)　(h)　(i)　(j)　(k)　(l)　(m)

Fig. 3 Visual Comparisons of extraction results for GeoEye-1 dataset. From left to right are: (a) raw image, (b) ground truth, (c) DSR, (d) GLSA, (e)Grad-CAM, (f) Grad-CAM++, (g) HWSL, (h) Itti, (i) MFF, (j) RBD, (k) SR, (l) SUN, (m) Our proposal.

According to MoCo [9], we introduced InfoNCE loss $L_q$ to measure the distance of positive and negative features:

$$L_{q1} = -\log \frac{\exp(q \cdot k_+ / \tau)}{\sum_0^K \exp(q \cdot k_i / \tau)} \quad (6)$$

where $\tau$ is a hyper-parameter, $k_i$ is a set of encoded samples and $k_+$ is the single key that $q$ matches. Related to our paper, $q \cdot k_+$ is equal to $l_{pos}$, and $q \cdot k_i$ is $l_{pos}$ when $i = 0$, otherwise is $l_{neg}$. Hence, we can employ InfoNCE to optimize our model. Similarly, we can obtain $L_{q2}$ by network prediction $\tilde{Y}$ and $up_{2f}$.

### 2.3 Overall Loss Function

The overall loss function includes two parts. The first part is the cross-entropy loss $loss_{ce}$ based on the provided pseudo labels $Y^c$ from our first part and the corresponding probability $P^c$:

$$Loss_{ce} = -\sum_c Y^c \cdot \log(P^c) \quad (7)$$

The second part is contrastive loss $L_{q1}$ and $L_{q2}$. The loss function is formulated as follows:

$$Loss_{joint} = Loss_{ce} + L_{q1} + L_{q2} \quad (8)$$

## 3. EXPERIMENTS

### 3.1 Dataset and Implemented Details

To validate the efficiency of the proposed method, we evaluate the performance on the GeoEye-1 dataset, which composes of 259 images with a resolution of 0.46m. We first resize the raw images to 256×256×3 before feeding them to our approach. In the dataset, images with residential areas are categorized as foreground images otherwise it will be background images. There are 129 foreground images and 151 background images. We train our model on the PC equipped with an Intel Core i9-10940X and a GPU NVIDIA Geforce RTX 3090.

As for classification CNN, our model is trained with cross-entropy loss function. The model is optimized by stochastic gradient descent with a batch size of 16 examples, a momentum of 0.9, and a weight decay of 0.0001. As for ROI extraction network, Adam optimizer is applied to train the network. The initial learning rate is set to be 5e-5 and multiplies 0.5 every 20 epochs. Due to GPU limits, we set batch size to 2 for all experiments. The train ends after 100 epochs.

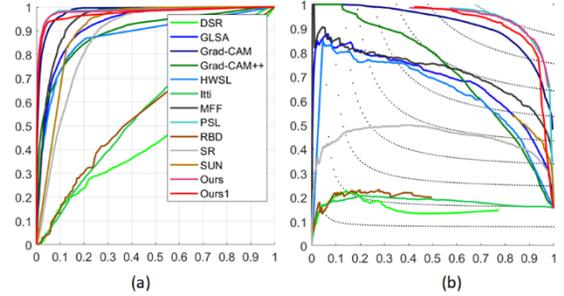

Fig. 4 Evaluation metrics for GeoEye-1 dataset: (a) PR curve, (b) ROC curve.

Tab. 1 Quantitative Comparisons for GeoEye-1 dataset

| Method_name | AC | AUC | Precision | Recall | F |
|---|---|---|---|---|---|
| DSR | 0.771 | 0.469 | 0.206 | 0.185 | 0.190 |
| GLSA | 0.805 | 0.926 | 0.448 | 0.921 | 0.502 |
| Grad-CAM | 0.950 | **0.982** | 0.845 | 0.836 | 0.840 |
| Grad-CAM++ | 0.851 | 0.896 | 0.574 | 0.792 | 0.590 |
| HWSL | 0.883 | 0.878 | 0.615 | 0.689 | 0.620 |
| Itti | 0.646 | 0.572 | 0.181 | 0.379 | 0.204 |
| MFF | 0.887 | 0.948 | 0.587 | 0.899 | 0.633 |
| RBD | 0.791 | 0.572 | 0.238 | 0.109 | 0.177 |
| SR | 0.783 | 0.875 | 0.408 | 0.898 | 0.462 |
| SUN | 0.646 | 0.911 | 0.323 | **0.993** | 0.378 |
| Ours | **0.964** | 0.981 | **0.896** | 0.877 | **0.890** |

### 3.2 Results and Comparisons

We compared the proposed method with ten state-of-art object segmentation methods proposed recently. In the following, both visual performances and quantitative analysis will be presented.

*Visual Comparison:* the mapping results are displayed in the Fig. 3. As observed, traditional methods cannot provide satisfactory results. For example, the extraction results of GLSA[5], Grad-CAM++[10], Itti[11] and SR[12] cannot capture complicated boundaries, and the non-ROIs with complex textures are detected. DSR[13], RBD[14], SUN[15] generate maps containing many nonresidential results, such as rivers and roads. Although Grad-CAM, HWSL[4] and MFF[16] produce better segmentation results, the results show shortcomings in boundary maintenance. As shown in column (m), our proposed method can provide a distinct and integrated contour.

*Quantitative Analysis:* We employ several measurements including overall accuracy (AC), Precision, Recall, F-

measure, area under the curve (AUC), the receiver operating characteristic (ROC) curve and precision–recall (PR) curve to evaluate our proposal. Fig.4 shows the comparison of the ROC curves and PR curves. The quantitative results are reported in Table 1. For PR curves, OTSU thresholding is utilized to segment the binary maps. For F-measure, we choose $\beta^2 = 0.3$ to evaluate the proposed method.

Among these methods, our proposal can achieve the highest overall accuracy. Grad-CAM and our method both show a generally high value of AUC. But its values of precision, recall and F score are lower than our method. Taking overall effectiveness and accuracy into account, our proposed method gives a more promising result.

### 3.3 Ablation Studies

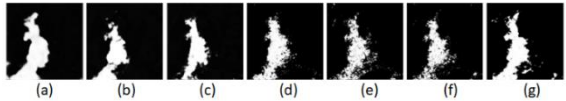

Fig. 5 The visual comparisons in ablation study: (a) without contrastive loss, (b) up1, (c) up2, (d) up3, (e) up2+up3, (f) up1+up3, (g) up1+up2(our proposal).

*Contrast learning:* In Fig. 5, column (a) shows the extraction result without contrast learning and column (g) is our proposal. We can observe that contrastive learning method can bring a better maintenance to our extraction results.

*The selection of upsampling layers:* Fig. 5 columns (b)-(g) display the visual performances when we utilize features from different upsampling layers in UNet architecture. We found that the results of deeper layers have a better boundary maintenance but worse ROI integrity. Since the proposal integrates features from different scales, the result of our proposal shows both a good boundary and region integrity.

### 4. CONCLUSION

In this paper, we propose a ROI extraction framework that combines weakly supervised learning method and contrastive learning for RSIs. First, we propose a pseudo label generation based on multiscale Grad-CAM in a weakly supervised manner, which reduces the dependence on large-scale and high-quality labeled data. Second, inspired by contrastive learning strategy, we construct a contrastive loss to address the mistakes in generated pseudo labels. In summary, our established model can achieve high accuracy and efficiency performances. In the future, we will explore the possibility of employing our model to multi-label weakly supervised tasks.

### ACKNOWLEDGEMENT

This work was supported by Chinese natural science foundation projects(62101052).

### 5. REFERENCES

[1] E. Shelhamer, J. Long, and T. Darrell, "Fully Convolutional Networks for Semantic Segmentation," *IEEE Transactions on Pattern Analysis and Machine Intelligence,* vol. 39, no. 4, pp. 640-651, 2017.

[2] O. Ronneberger, P. Fischer, and T. Brox, "U-net: Convolutional networks for biomedical image segmentation," *Lecture Notes in Computer Science (including subseries Lecture Notes in Artificial Intelligence and Lecture Notes in Bioinformatics).* pp. 234-241.

[3] J. Zhang, X. Jia, and J. Hu, "SP-RAN: Self-Paced Residual Aggregated Network for Solar Panel Mapping in Weakly Labeled Aerial Images," *IEEE Transactions on Geoscience and Remote Sensing,* vol. 60, 2022.

[4] L. Zhang, J. Ma, X. Lv, and D. Chen, "Hierarchical Weakly Supervised Learning for Residential Area Semantic Segmentation in Remote Sensing Images," *IEEE Geoscience and Remote Sensing Letters,* vol. 17, no. 1, pp. 117-121, 2020.

[5] L. Zhang, A. Li, Z. Zhang, and K. Yang, "Global and local saliency analysis for the extraction of residential areas in high-spatial-resolution remote sensing image," *IEEE Transactions on Geoscience and Remote Sensing,* vol. 54, no. 7, pp. 3750-3763, 2016.

[6] J. M. Murphy, and M. Maggioni, "Unsupervised Discriminative Dimension Reduction for Hyperspectral Chemical Plume Segmentation," *International Geoscience and Remote Sensing Symposium (IGARSS).* pp. 3828-3831.

[7] Z. Chu, T. Tian, R. Feng, and L. Wang, "Sea-land segmentation with res-unet and fully connected CRF," *International Geoscience and Remote Sensing Symposium (IGARSS).* pp. 3840-3843.

[8] R. R. Selvaraju, M. Cogswell, A. Das, R. Vedantam, D. Parikh, and D. Batra, "Grad-CAM: Visual Explanations from Deep Networks via Gradient-Based Localization," *International Journal of Computer Vision,* vol. 128, no. 2, pp. 336-359, 2020.

[9] K. He, H. Fan, Y. Wu, S. Xie, and R. Girshick, "Momentum Contrast for Unsupervised Visual Representation Learning," *Proceedings of the IEEE Computer Society Conference on Computer Vision and Pattern Recognition.* pp. 9726-9735.

[10] A. Chattopadhay, A. Sarkar, P. Howlader, and V. N. Balasubramanian, "Grad-CAM++: Generalized gradient-based visual explanations for deep convolutional networks," *Proceedings - 2018 IEEE Winter Conference on Applications of Computer Vision, WACV 2018.* pp. 839-847.

[11] L. Itti, C. Koch, and E. Niebur, "Model of saliency-based visual attention for rapid scene analysis," *IEEE Transactions on Pattern Analysis and Machine Intelligence,* vol. 20, no. 11, pp. 1254-1259, 1998.

[12] X. Hou, and L. Zhang, "Saliency detection: A spectral residual approach," *Proceedings of the IEEE Computer Society Conference on Computer Vision and Pattern Recognition.*

[13] X. Li, H. Lu, L. Zhang, X. Ruan, and M.-H. Yang, "Saliency detection via dense and sparse reconstruction," *Proceedings of the IEEE International Conference on Computer Vision.* pp. 2976-2983.

[14] W. Zhu, S. Liang, Y. Wei, and J. Sun, "Saliency optimization from robust background detection," *Proceedings of the IEEE Computer Society Conference on Computer Vision and Pattern Recognition.* pp. 2814-2821.

[15] L. Zhang, M. H. Tong, T. K. Marks, H. Shan, and G. W. Cottrell, "SUN: A Bayesian framework for saliency using natural statistics," *Journal of vision,* vol. 8, no. 7, pp. 32-32, 2008.

[16] L. Zhang, K. Yang, and H. Li, "Regions of interest detection in panchromatic remote sensing images based on multiscale feature fusion," *IEEE Journal of Selected Topics in Applied Earth Observations and Remote Sensing,* vol. 7, no. 12, pp. 4704-4716, 2014.